\documentclass[10pt, logo, twocolumn, copyright]{nvidiatechreport}

\usepackage[utf8]{inputenc}
\usepackage[T1]{fontenc}

\usepackage{amsmath}
\usepackage{amsfonts}
\usepackage{nicefrac}

\usepackage{microtype}
\usepackage{xspace}

\usepackage[dvipsnames]{xcolor}
\usepackage{colortbl}

\usepackage{graphicx}
\usepackage{subcaption}
\usepackage{wrapfig}
\usepackage{float}
\usepackage{dblfloatfix}

\usepackage{booktabs}
\usepackage{multirow}
\usepackage{makecell}
\usepackage{adjustbox}
\usepackage{tabularx}
\usepackage{arydshln}
\usepackage{hhline}
\usepackage{array}
\usepackage{diagbox}
\usepackage{hyperref}
\usepackage{enumitem}
\usepackage{multicol}

\usepackage{mdframed}
\usepackage[most]{tcolorbox}

\usepackage[ruled,vlined,linesnumbered]{algorithm2e}

\usepackage{footmisc}

\usepackage[numbers]{natbib}

\usepackage{tabto}
\def\Snospace~{Section }

\usepackage{url}
\usepackage{hyperref}

\usepackage{amsmath,amsfonts,bm}

\def\eqref#1{equation~\ref{#1}}

\def\1{\bm{1}}

\DeclareMathAlphabet{\mathsfit}{\encodingdefault}{\sfdefault}{m}{sl}
\SetMathAlphabet{\mathsfit}{bold}{\encodingdefault}{\sfdefault}{bx}{n}

\definecolor{Note_color}{rgb}{0.0, 0.0, 0.0}

\definecolor{lightyellow}{RGB}{255, 255, 204}
\definecolor{lightred}{RGB}{255, 204, 204}
\definecolor{lightgreen}{RGB}{204, 255, 204}

\definecolor{darkblue}{rgb}{0.3, 0.5, 0.9}
\definecolor{projectgreen}{HTML}{76B900}
\definecolor{lightblue}{RGB}{220,230,255}
\definecolor{selfblue}{RGB}{65,105,225} 
\definecolor{Gray}{gray}{0.93}
\hypersetup{colorlinks=true, citecolor=darkblue, linkcolor=darkblue, urlcolor=darkblue}

\newcommand{\sys}{EdgeWise\xspace}
\usepackage{xcolor}
\definecolor{maskorange}{HTML}{ffe6cc}
\definecolor{maskblue}{HTML}{d4e1f5}
\definecolor{maskred}{HTML}{ffcccc}
\definecolor{maskgreen}{HTML}{d5e7d4}
\definecolor{diff_green}{HTML}{7cb342}
\definecolor{nldgreen}{HTML}{E8F5E9}

\newcommand{\teaserfigure}{}

\usepackage{pifont}
\newcommand\mynuma[1]{\ifcase#1 \or \ding{172}\or \ding{173}\or
  \ding{174}\or \ding{175}\or \ding{176}\or \ding{177}
  \or \ding{178}\or \ding{179}\or \ding{180}\or \ding{181}\else *\fi\relax}

\newcommand\mynumb[1]{\ifcase#1 \or \ding{182}\or \ding{183}\or
  \ding{184}\or \ding{185}\or \ding{186}\or \ding{187}
  \or \ding{188}\or \ding{189}\or \ding{190}\or \ding{191}\else *\fi\relax}

\title{EdgeWise: Bringing Long-Horizon Agents to the Edge via Selective Cloud Delegation}

\author{Zeru Shi$^{1\dagger*}$, Jindong Jiang$^{*}$, Ryo Hachiuma, Yonggan Fu, Byung-Kwan Lee, Peter Belcak, Saurav Muralidharan, Ayush Agarwal, Ruixiang Tang$^{1}$, Ambrish Dantrey, Pavlo Molchanov, Anjul Patney, Subhashree Radhakrishnan}

\correspondingauthor{X}
\projectpage{https://research.nvidia.com/labs/lpr/edge-wise/}

\begin{abstract}
Complex and long-horizon agent tasks often rely on frontier models deployed in the cloud to achieve strong performance. However, using these models for every step of a long-horizon trajectory incurs substantial inference cost, even though not all steps may require the same level of model capability. In this paper, we find that allocating the LLM to only a small subset of key steps can achieve performance comparable to full LLM execution while significantly reducing inference cost. Motivated by this observation, we propose \textbf{\textit{\sys}}, an edge-cloud collaboration framework that deploys an SLM on the local device as the primary solver and selectively delegates critical steps to a cloud LLM when stronger model capacity is needed. To enable autonomous delegation without additional models at inference time, we introduce a step-level RL method that jointly optimizes task solving and delegation, with fine-grained credit assignment for individual delegation decisions. We also introduce an external judge-based delegation strategy and use it to systematically investigate key design choices in edge-cloud collaboration, including when to delegate, how delegation decisions should be made, what context should be shared, and how frequently the cloud LLM should be invoked. Experiments show that \sys matches or outperforms pure-LLM execution while reducing inference cost up to 62\% on SWE-Rebench and 34\% on Terminal-Bench 2.1. We further deploy \sys on a local workstation, demonstrating comparable cost savings in a practical edge-cloud setting.

\end{abstract}

\makeatletter
\renewcommand{\maketitle}{%
  \twocolumn[%
    \begin{adjustwidth}{0pt}{24pt}%
      \begin{flushleft}%
        {\raggedright \titlefont \@title\par}%
        \vskip11pt%
        {\raggedright \@author\par}%
        \vskip20pt%
      \end{flushleft}%
    \end{adjustwidth}%
    \abscontent
    \vskip16pt
    \teaserfigure
    \vskip20pt%
  ]%
  \thispagestyle{firststyle}%
}
\makeatother

\renewcommand{\teaserfigure}{%
  \begin{center}
    \includegraphics[width=\textwidth]{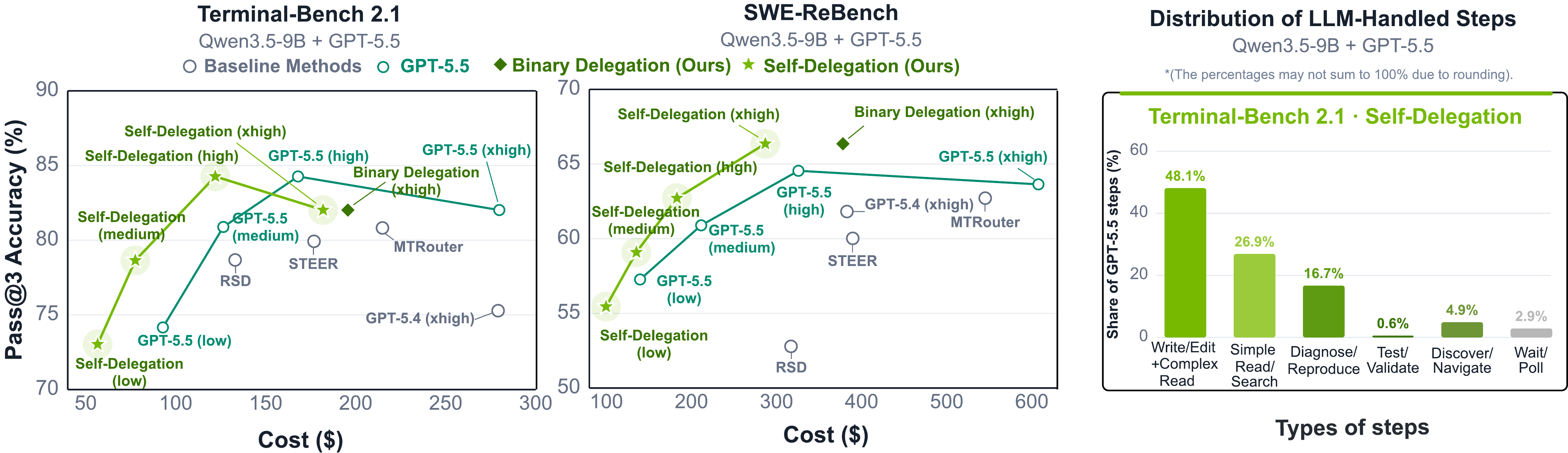}
    \captionof{figure}{The two left figures illustrate the trade-off between Pass@3 performance and inference cost for our method and the baselines across different settings and efforts. The LLM used for the comparison method is GPT-5.5. The right figure further analyzes the delegation behavior of our method by showing the proportions of different types of agent steps that are delegated to the LLM.}
    \label{fig:first}
    \vspace{-15pt}
  \end{center}
}

\begin{document}
\maketitle

\section{Introduction}
\label{sec:intro}

\begin{figure*}[t]
    \centering
    \includegraphics[width=\textwidth]{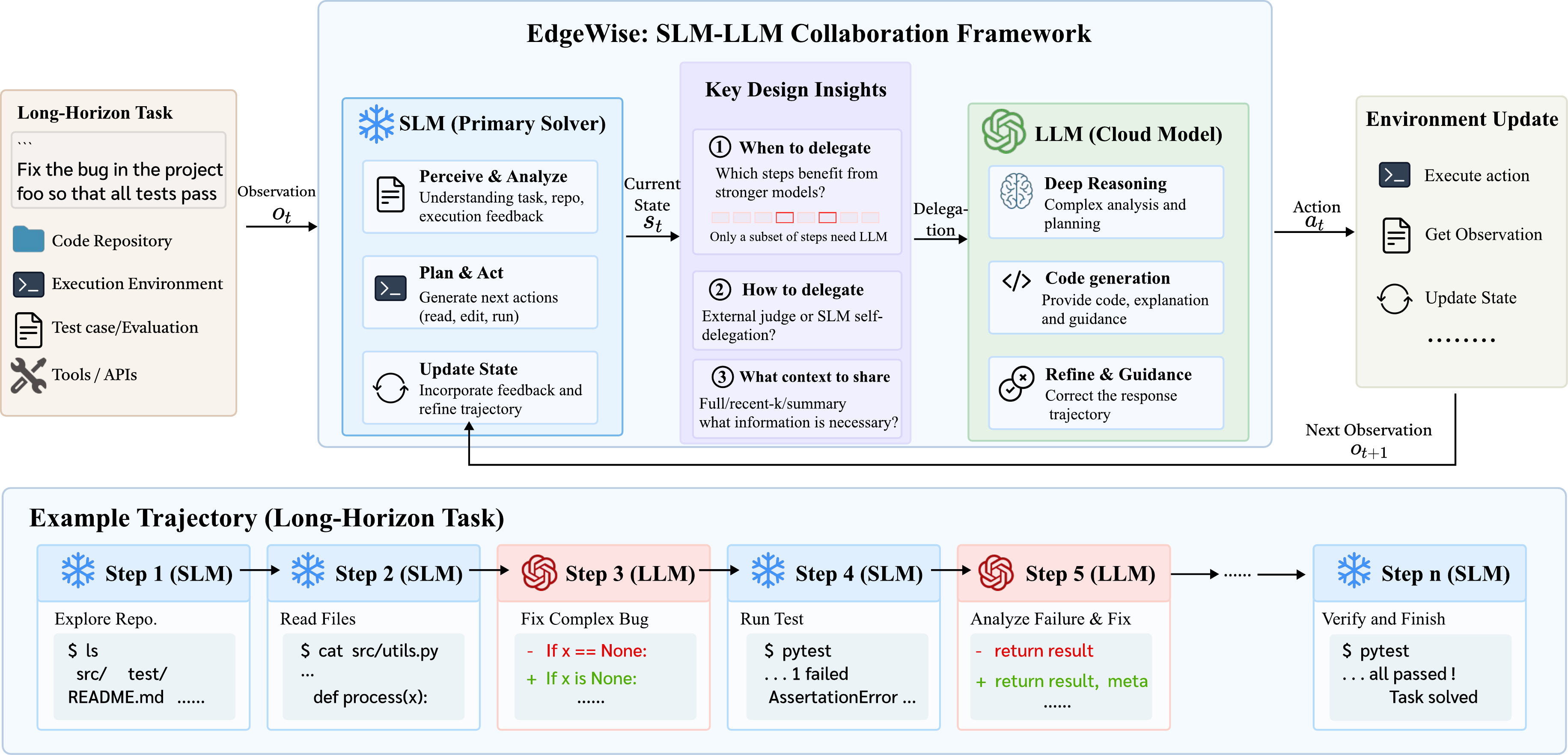}
    \caption{The upper panel illustrates the overall pipeline of \sys, while the lower panel provides an example of how \sys operates during task execution.}
    \label{fig:judge-context}
    \vspace{-15pt}
\end{figure*}
LLM-based agents have achieved strong performance on increasingly complex long-horizon tasks~\citep{team2026kimi,chandiramani2026nemotron,GPT-5.6}. However, frontier LLMs require substantial computation and are typically served through cloud infrastructure, making it costly to rely on them throughout long-horizon task execution. Importantly, model capability requirements can vary across intermediate steps, and many simpler steps can be handled by smaller models deployed locally. Recent works have explored allocating queries of different difficulty levels to models with varying capabilities.~\citep{panda2025adaptive,maurya2025selectllm,ramirez2024optimising,aggarwal2024automix}. Beyond single-query routing, Router-R1~\citep{zhang2026router} and MERA~\citep{yao2026mera} extend model selection to multi-step settings, dynamically determining which model should handle intermediate invocations as the task progresses. However, existing methods still have critical limitations. (1) \textbf{Limited Long-Horizon Evaluation}: As noted by TwinRouterBench~\citep{yang2026twinrouterbench}, most existing methods are evaluated on single-query reasoning or multi-hop QA, leaving real-world long-horizon tasks such as coding and deep research largely unexplored. (2) \textbf{Model-Pool Dependency}: Existing approaches typically rely on additional decision models trained over predefined model pools, making their decisions dependent on specific model capabilities and potentially less reliable as the model pool evolves.

Motivated by these limitations, as well as recent advances in SLM capabilities and edge computing hardware that make it increasingly practical to deploy capable SLMs directly on local devices, such as the NVIDIA DGX Spark\footnote{\url{https://build.nvidia.com/spark}}, we consider an edge--cloud setting where a locally deployed SLM serves as the primary solver for long-horizon tasks and selectively leverages a stronger cloud LLM when needed. Based on this setting, we propose \textbf{\sys}, a step-level edge--cloud collaboration framework in which the local SLM continuously interacts with the environment and selectively delegates critical steps to the cloud LLM. We explore two delegation mechanisms: an external judge-based strategy that dynamically determines when the cloud LLM is needed, and a self-delegation strategy that integrates this capability directly into the SLM through step-level RL with fine-grained credit assignment. Beyond these mechanisms, we systematically analyze the key design factors of edge--cloud collaboration, including when to delegate, what context should be shared, and how delegation frequency should be controlled. Our analysis reveals that the benefit of the cloud LLM is concentrated in a small subset of critical steps, enabling \sys to reduce cloud-model usage while preserving the performance of full LLM execution. Experimental results show that our methods, particularly RL-based self-delegation, maintain competitive performance while reducing inference cost by an average of 46\% across models and benchmarks. We further validate \sys in a practical edge--cloud deployment, where the locally deployed SLM achieves comparable performance and cost savings. \autoref{fig:first} shows the Pass@3--cost trade-off and demonstrates that our method selectively delegates harder steps to the LLM, with further analysis in~\autoref{distribution}. In summary, our main contributions are as follows:
\begin{itemize}[noitemsep, topsep=0pt]
\item We propose \textbf{\textit{\sys}}, a step-level edge--cloud collaboration framework for long-horizon agents, where an SLM deployed on the local device serves as the primary solver and selectively leverages a stronger cloud LLM when needed.

\item We systematically investigate the \textbf{key design factors of \sys}, including when the cloud LLM is needed, how delegation decisions should be made, what context should be shared between local and cloud models, and how frequently the cloud LLM should be invoked. Our analysis reveals that the benefit of the cloud LLM is concentrated in a small subset of critical steps.

\item Within \sys, we introduce two cloud delegation mechanisms: external-judge-based delegation and self-delegation. In particular, our \textbf{step-level RL method with fine-grained credit assignment} enables the local SLM to jointly learn task solving and when to leverage the cloud LLM, without an additional decision model at inference time.

\item Extensive experiments on long-horizon coding-agent benchmarks demonstrate that our self-delegation method maintains competitive task performance while reducing inference cost by an average of 46\% across different LLMs and benchmarks. We further conduct a real-world deployment evaluation, demonstrating the practical feasibility of our edge--cloud collaboration framework for long-horizon agents.
\end{itemize}

\begin{figure*}[t]
    \centering
    \includegraphics[width=\textwidth]{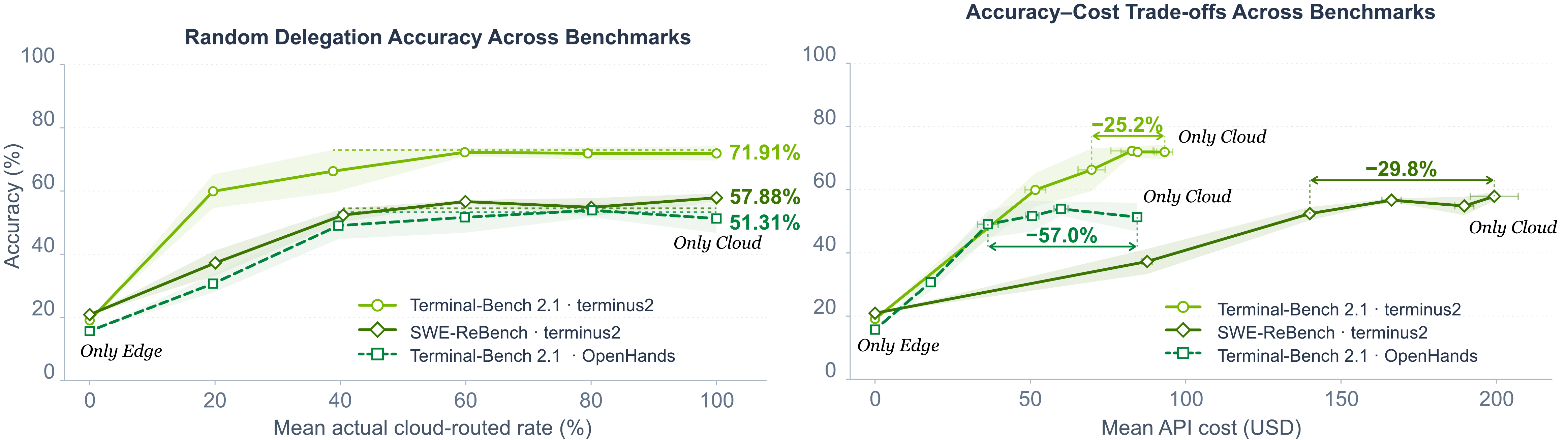}
    \caption{The leftmost and rightmost points represent Qwen3.5-9B-only and GPT-5.5-only, respectively, with four intermediate LLM usage ratios. The left figure shows accuracy as the LLM ratio increases, while the right reports the corresponding average cost over three runs. Each point is averaged over three runs, with shaded regions indicating standard deviation. This figure shows that, with a cloud usage rate of only around 40\%, delegating the right steps to the cloud model achieves performance comparable to the vanilla model while reducing cost.} 
    \label{fig:random-selection}
\end{figure*}
\section{Observation: Delegating Only Key Steps Matches Large-Model Performance}
\label{observation}

Long-horizon coding tasks typically involve a sequence of heterogeneous steps, including problem understanding, root-cause localization, code editing, testing, and result inspection. These steps differ in complexity and may therefore require different levels of model capability. Nevertheless, existing agent harnesses typically rely on a large model throughout the entire trajectory, allocating the same level of model capability to every step. This raises a natural question: \textit{Does every step in a long-horizon trajectory require a large model?} To investigate this question, we first conduct a random delegation experiment. We use the SLM as the primary solver and randomly delegate a subset of steps to the LLM. Before each step, a delegation decision is independently sampled according to a predefined probability. If selected, the LLM executes the current step, after which control returns to the SLM for subsequent execution; otherwise, the SLM handles the step locally. This setup allows us to examine how task performance changes when the LLM participates in only a subset of steps, and whether the placement of these LLM calls affects the overall performance.

In our experiments, we use Qwen3.5-9B~\citep{qwen3.5} as the local model and GPT-5.5~\citep{openai2026gpt55} with xhigh reasoning effort as the cloud model, with a maximum of 80 turns per trajectory. We follow the default settings of each benchmark, including the original execution-time constraints, and evaluate on Terminal-Bench 2.1 and SWE-Rebench using two agent harnesses, Terminus2 and OpenHands~\citep{openhands}. To investigate whether the allocation of cloud intelligence matters, we randomly select a fixed proportion of steps to be handled by the cloud model. As shown in~\autoref{fig:random-selection}, when the cloud-model ratio reaches 40\%, performance exhibits substantial variation across different random allocations. Notably, some allocations achieve performance comparable to using the cloud model throughout the entire trajectory, while reducing inference cost by at least 25\%. This suggests that \textbf{a relatively small number of well-placed cloud-model calls can be sufficient for strong task performance, with the local model effectively handling the remaining steps.} Overall, these results indicate that effective edge SLM--cloud LLM collaboration depends not only on \textit{how much} stronger model capacity is used, but more importantly on \textit{where} it is allocated along the trajectory.

\section{EdgeWise: A Framework for SLM--LLM Collaboration in Problem Solving}

Based on this finding, we introduce \textbf{\sys}, an edge SLM and cloud LLM collaboration framework that uses the SLM as the primary solver and selectively delegates challenging steps to the LLM when stronger capability is needed. In this section, we systematically investigate the key design choices of such collaboration, including when and how to delegate, with the goal of maintaining strong task performance while reducing LLM usage and inference cost.

\subsection{When Should the SLM Delegate to the LLM?}
\label{when}

\begin{figure*}[t]
    \centering
    \includegraphics[width=\textwidth]{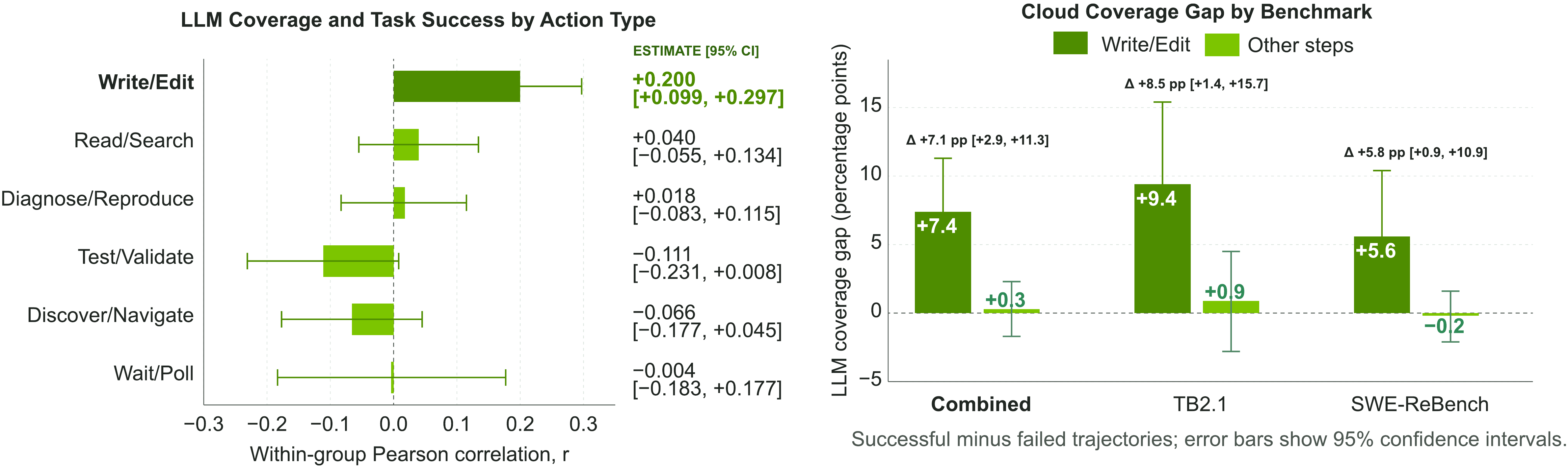}
    \caption{Left: Correlation between cloud coverage and task success across action types. Right: Cloud coverage differences between successful and failed trajectories. Error bars denote 95\% confidence intervals. Higher coverage of Write/Edit steps is consistently associated with task success.}
    \label{fig:step_type_ana}
    \vspace{-5pt}
\end{figure*}

In \autoref{observation}, we observe that delegating only a subset of steps to the LLM can substantially reduce inference cost, with some random allocations achieving performance comparable to full LLM execution. However, at a 40\% delegation ratio, performance varies considerably across random allocations, suggesting that \textbf{where to delegate} is critical under a limited budget. This motivates the first key question in designing \sys: \textbf{Which steps along the trajectory should be delegated to the large model?} To investigate this question, we analyze 2.4k coding-agent trajectories and categorize each step into six action types: \textit{Write/Edit}, \textit{Read/Search}, \textit{Diagnose/Reproduce}, \textit{Test/Validate}, \textit{Discover/Navigate}, and \textit{Wait/Poll}. For each type, we measure its cloud coverage, defined as the fraction of steps delegated to the cloud model. To control for task difficulty and delegation budget, we group trajectories by benchmark, task, and random-selection budget, and center both cloud coverage and binary reward within each group. We then pool the centered observations and compute the Pearson correlation:$
r_a =
\frac{\sum_i \widetilde{C}_{a,i}\widetilde{R}_i}
{\sqrt{\sum_i \widetilde{C}_{a,i}^{2}}\sqrt{\sum_i \widetilde{R}_i^{2}}},
$
where $\widetilde{C}_{a,i}$ and $\widetilde{R}_i$ denote the centered cloud coverage of action type $a$ and task reward, respectively. This measures whether greater cloud coverage of a particular action type is associated with higher task success under the same task and delegation budget. We estimate 95\% confidence intervals using cluster bootstrap over task-and-budget groups.

As shown in the left panel of \autoref{fig:step_type_ana}, \textit{Write/Edit} is the action type whose cloud coverage is most strongly associated with task success, with a correlation of $0.200$ and a 95\% confidence interval of $[0.099, 0.297]$. \textit{Read/Search} shows a weaker positive association, while the remaining action types have smaller correlations with confidence intervals overlapping zero. The right panel further supports this finding: successful trajectories allocate more cloud coverage to \textit{Write/Edit}, with gaps of $9.4$ percentage points on Terminal-Bench 2.1 and $5.6$ on SWE-ReBench, while differences for other action types remain small. Together, these results identify \textbf{Write/Edit} as the action type most strongly associated with task success, suggesting that under a limited delegation budget, performance depends more on delegating critical steps than on uniformly increasing cloud-model usage.

\subsection{Who Should Decide Whether to Delegate?}
\label{who}

Our analysis suggests that different steps benefit differently from stronger model allocation, motivating a more selective delegation strategy. This raises a practical question: \textit{how should the delegation decision be made at each step?} We explore two  mechanisms. The first uses an external judge to dynamically determine whether the current step should be delegated, while the second integrates this capability directly into the SLM, enabling it to jointly learn task solving and delegation.

\subsubsection{Key-Step-Aware Delegation with an External Judge}
\label{external_judge}

\begin{figure}[htbp]
    \centering
    \includegraphics[width=0.5\textwidth]{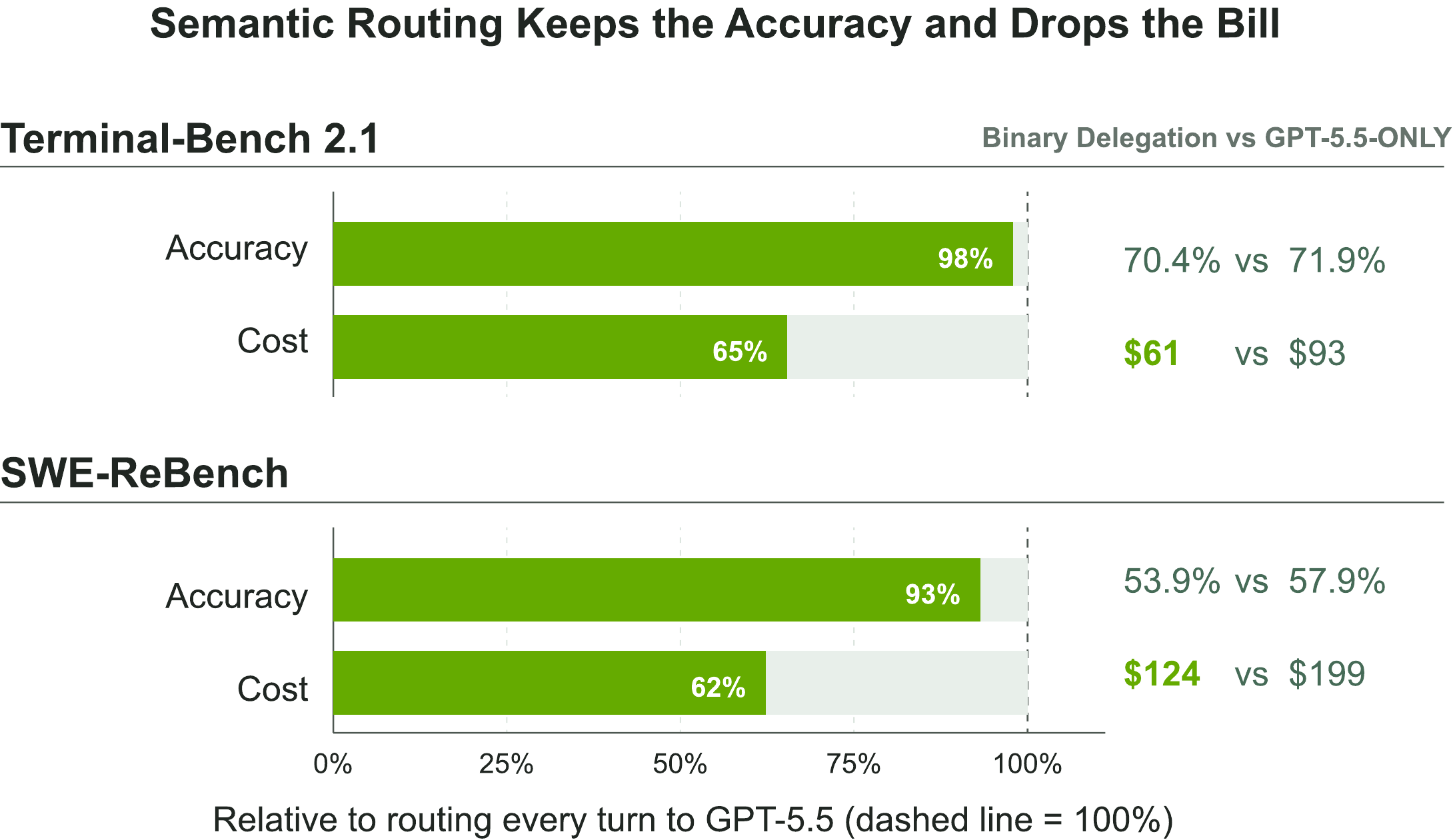}
    \captionof{figure}{Performance and API cost of our training-free key-step delegation method compared with the pure large-model baseline. All metrics are averaged over three independent runs.}
    \label{fig:traing-free-vs-pure-cloud}
    \vspace{-5pt}
\end{figure}

Building on the key-step analysis in~\autoref{when}, we design a key step-aware delegation mechanism that selectively delegates critical steps to the cloud LLM. Specifically, based on the patterns observed in Figure~\autoref{fig:step_type_ana}, we use Write/Edit and complex Read/Search operations as guidance for identifying key steps in our delegation mechanisms, and translate these patterns into delegation criteria. We then introduce an external judge to dynamically identify such key steps and determine whether the current step should be delegated to the cloud LLM throughout the trajectory. Given the trajectory history $h_t$ at step $t$, the judge determines whether the next step exhibits the characteristics of a key step and makes a binary delegation decision: $a_t = J(h_t), a_t \in \{\text{Local}, \text{Cloud}\}.$
If the current state is identified as requiring stronger model, the next step is delegated to the cloud model; otherwise, the local model continues execution. After each step, the judge reevaluates the updated trajectory, allowing key-step identification and model allocation to adapt dynamically as the task progresses. We instantiate GPT-5.4-mini as the external judge and incorporate the key-step patterns identified in~\autoref{fig:step_type_ana} into its delegation criteria. As shown in~\autoref{fig:traing-free-vs-pure-cloud}, this key step aware delegation strategy preserves most of the performance of the pure cloud-model baseline while reducing cloud-model usage. These results validate that the identified key-step patterns can effectively guide delegation, further supporting our finding in~\autoref{when} that delegating Write/Edit and complex Read/Search steps to the LLM benefits SLM task solving.
We therefore use this mechanism as the training-free variant of \sys and further investigate its design choices, including LLM context (\autoref{what_llm}), delegation context (\autoref{what_judge}), and delegation frequency (\autoref{how_fre}).

\subsubsection{Training the SLM for Self-Delegation}
\label{rl_method}

In the previous section, we use an external judge to determine whether each step should be handled by the SLM or delegated to the LLM. To eliminate the need for an additional model, we formulate self-delegation as a joint task-solving and delegation problem, where the SLM learns to solve the task while selectively invoking the cloud LLM when additional model capability is useful. We adopt a two-stage training pipeline: SFT first teaches the SLM the delegation format, followed by RL that encourages efficient task solving while learning to delegate only critical steps to the LLM.

\textbf{Stage1: SFT training on based model.} Specifically, we use GPT-5.5 to generate task-solving trajectories. At each step, GPT-5.5 additionally considers whether the step would require delegation if executed by a smaller model, and generates the corresponding delegation label. Importantly, these labels are not intended to indicate that a step \textit{must} be delegated to the LLM. Instead, they serve as format supervision, enabling the SLM to learn delegation as a valid action. Given the constructed dataset $\mathcal{D}_{\mathrm{SFT}}=\{(x_i,y_i)\}_{i=1}^{N}$, we optimize the standard language modeling objective: $\mathcal{L}_{\mathrm{SFT}} = -\sum_{(x,y)\in\mathcal{D}_{\mathrm{SFT}}} \log p_{\theta}(y\mid x),$ where $y$ contains both the task-solving trajectory and the corresponding delegation actions.

\textbf{Stage 2: RL Training for Self-Delegation.}
After the first-stage training, we observe that although the SLM learns how to delegate to the LLM, it does not yet learn how to allocate cloud-model usage effectively across different steps, leading to suboptimal delegation behavior~\autoref{call_time} and degraded task performance~\autoref{sft-rl}. To address this issue, we introduce an RL-based training method that jointly encourages successful task completion and selective allocation of LLM calls to critical steps, as identified in~\autoref{when}. During training, we follow the output format introduced in the SFT stage. At each step, the SLM first decides whether to delegate to the LLM. For local execution, the SLM generates the analysis and action itself; otherwise, the LLM generates the subsequent output. Since the delegated output is generated by the LLM, these tokens are excluded from the policy loss. Our trajectory-level reward combines an outcome reward for task correctness with a cloud-usage regularization term:
\begin{equation}
R_i = V_i + \mathbb{I}(V_i=1)\frac{1}{1+\lambda C_i},
\end{equation} 
where $V_i$ indicates whether the final answer is correct and $C_i$ is the number of LLM calls. $\lambda$ is a hyperparameter determined by the typical number of cloud calls per trajectory. A larger number of cloud calls requires a smaller $\lambda$, and vice versa, to maintain an appropriate scale for the cost reward. In our experiments, we set $\lambda=0.1$ based on the cloud call statistics of the training data.  The second term favors successful trajectories that use fewer cloud invocations, serving as a regularizer on overall cloud usage rather than directly optimizing monetary inference cost. Importantly, this trajectory-level signal only controls the overall amount of delegation and does not specify which steps should receive the limited cloud-model capacity. To explicitly learn \textit{where} to allocate LLM calls, we introduce step-level credit assignment for individual delegation decisions. In standard GRPO, tokens within a sampled response typically share the same trajectory-level advantage, making it difficult to distinguish critical delegation decisions from routine ones. Based on the criteria derived in~\autoref{when}, we use GPT-5.4-mini as a lightweight critic model. Given the full rollout trajectory, the critic model evaluates each delegation decision according to the key-step criteria defined in~\autoref{when}. Decisions consistent with the critic model's assessment receive a positive advantage, while unnecessary delegation or missed delegation receives a negative advantage. No additional advantage is assigned when local execution is sufficient. In this way, the step-level signal does not simply encourage fewer delegations; instead, it encourages SLM to concentrate cloud-model usage on steps where delegation is most valuable. For delegation tokens at step $i$, its final advantage is
\begin{equation}
A_t = A_{\text{traj}} +
\beta \frac{N_{\text{trajectory}}}{N_{\text{route}}}
A_i^{\mathrm{del}},
\end{equation}
where $A_{\text{traj}}$ is the trajectory-level GRPO advantage, $N_{\text{trajectory}}$ is the number of training tokens in the trajectory, $A_i^{\mathrm{del}}$ is the token level advantage on delegation tokens and $N_{\text{route}}$ is the number of delegation tokens. The scaling factor prevents the step-level signal from being diluted when delegation tokens are sparse, while $\beta$ controls its relative contribution to the policy loss. We evaluate the effectiveness of this step-level credit assignment in~\autoref{token_adv_good} and further compare the SFT and RL models in~\autoref{sft-rl}. We can see our method not only reduces the number of delegations but also allocates cloud model capacity more selectively to critical steps. We show the agreement between SLM and critic model in~\autoref{learning_curve}.

\subsection{What Context Should Be Provided for Collaboration?}
\label{context_collaboration}

Effective SLM--LLM collaboration depends not only on when cloud model is introduced, but also on what information is available to different components when making their decisions. In \sys, context serves two distinct purposes. The LLM requires sufficient information to understand the current task state and effectively perform a delegated step, while the judge requires appropriate information to determine whether delegation is necessary in the first place. We therefore investigate the context provided to these two components separately and we empirically compare these configurations in~\autoref{context_exp}.

\textbf{What Information Should Be Passed to the LLM?}
\label{what_llm}
When a step is delegated, the LLM requires sufficient context to understand the current task state, while longer context also increases inference cost. We therefore investigate five context configurations: \textit{(1) Full History}, which provides the complete preceding trajectory; \textit{(2) Previous Step}, which provides only the analysis, action, and observation from the preceding step; \textit{(3) Recent History}, which provides the most recent $k$ turns; \textit{(4) No SLM Reasoning}, which removes the SLM's analysis while retaining executed commands and observations; and \textit{(5) Relevant Steps}, where the judge selects previous steps most relevant to the upcoming delegated step. These configurations examine the trade-off between context completeness and inference cost.

\textbf{What Information Should Be Passed to the Judge?}
\label{what_judge}
Unlike the LLM, the judge only determines whether the next step should be delegated and may therefore require different contextual information. We consider three configurations for step $n$: \textit{(1) Full History}, containing steps $1$ to $n-1$; \textit{(2) Full History + Current Analysis}, which additionally includes the SLM's analysis for step $n$; and \textit{(3) Current Analysis Only}, which provides only the SLM's current analysis.

We empirically compare these configurations in~\autoref{context_exp} and adopt \textit{Full History} as the default context for the judge in \sys.

\subsection{How Frequently Should the LLM Be Invoked?}
\label{how_fre}

Beyond determining which steps should be delegated, another important design choice is how frequently the LLM should be invoked. More frequent delegation provides greater access to stronger model capabilities but also increases inference cost, whereas overly conservative delegation may prevent the LLM from intervening at critical steps. Therefore, \sys requires a mechanism to explicitly control LLM involvement and balance task performance against inference cost.

In~\autoref{external_judge}, the external judge makes a binary decision at each step, determining whether the next step should be handled by the SLM or delegated to the LLM. While effective for identifying potential delegation opportunities, binary decisions provide limited control over the overall delegation frequency. To address this limitation, we extend the judge to produce a continuous \textit{delegation score}. Given the trajectory history $h_t$ at step $t$, the judge produces $s_t = J(h_t)$, where $s_t$ represents the estimated necessity of invoking the LLM for the next step. The delegation decision is then determined by a predefined threshold $\tau$:$
a_t =
\begin{cases}
\text{Cloud}, & s_t > \tau, \\
\text{Local}, & s_t \leq \tau.
\end{cases}$ After each step, the judge reevaluates the updated trajectory and produces a new delegation score, allowing the decision to adapt as task execution progresses. By adjusting $\tau$, we can directly control the level of LLM involvement: a lower threshold results in more frequent delegation, while a higher threshold keeps more steps on the local SLM. This mechanism enables \sys to operate under different cloud budgets and provides a flexible way to explore the trade-off between task performance and inference cost.\autoref{cloud_budget} shows the relationship between the delegation threshold and the resulting cloud-call budget.

\section{Experiments}
\subsection{Experimental Setup}
For training, we use SWE-Smith with Qwen3.5-9B as the SLM and Kimi-K3 as the LLM. We use 2K examples for SFT, followed by RL training on 40 NVIDIA A100 80GB GPUs for 175 steps, with a batch size of 8 and 16 rollouts per example. We refer to the resulting trained checkpoint as EdgeWise-9B. We evaluate on Terminal-Bench 2.1 and SWE-ReBench-03 using the Terminus2 harness, limiting each trajectory to 80 turns while retaining the default benchmark settings. Learning curve and cloud usage are shown in~\autoref{learning_curve} and~\autoref{call_time}, respectively.

\subsection{Main Results of Our Framework}

\begin{table*}[t]
\centering
\caption{Performance and cost across benchmarks and baselines. Binary and Threshold Delegation costs include the external judge. \textbf{Bold} and \underline{underlined} denote the best and second-best results excluding vanilla LLM. The last rows report Self-Delegation changes relative to vanilla LLM.}
\label{tab:main_results}
\vspace{-8pt}
\setlength{\tabcolsep}{5.2pt}
\renewcommand{\arraystretch}{1.12}

\resizebox{\textwidth}{!}{
\begin{tabular}{lcccccccc}
\toprule

\multirow{3}{*}{\textbf{Method}}
& \multicolumn{4}{c}{\textbf{Qwen3.5-9B + GPT-5.5(xhigh)}}
& \multicolumn{4}{c}{\textbf{Qwen3.5-9B + Opus4.8(xhigh)}} \\

\cmidrule(lr){2-5}
\cmidrule(lr){6-9}

& \multicolumn{2}{c}{\textit{Pass@1}}
& \multicolumn{2}{c}{\textit{Pass@3}}
& \multicolumn{2}{c}{\textit{Pass@1}}
& \multicolumn{2}{c}{\textit{Pass@3}} \\

\cmidrule(lr){2-3}
\cmidrule(lr){4-5}
\cmidrule(lr){6-7}
\cmidrule(lr){8-9}

& Acc. (\%) $\uparrow$ & Cost (\$) $\downarrow$
& Acc. (\%) $\uparrow$ & Cost (\$) $\downarrow$
& Acc. (\%) $\uparrow$ & Cost (\$) $\downarrow$
& Acc. (\%) $\uparrow$ & Cost (\$) $\downarrow$ \\

\midrule

\multicolumn{9}{c}{\large\textit{\textbf{Terminal-Bench2.1}}} \\


Vanilla LLM
& 74.16 & 90.31
& 82.02 & 279.73
& 67.42 & 64.78
& 75.28 & 187.72 \\

Random Delegation
& 66.29 & 74.34
& 79.78 & 209.13
& 58.42 & 54.00
& 67.42 & 158.18 \\

RSD\citep{liao2025reward}
& 62.92 & \textbf{45.61}
& 78.65 & \textbf{133.25}
& 53.93 & \textbf{28.09}
& 61.80 & \textbf{82.39} \\

STEER\citep{lee2026confidence}
& 66.29 & 61.81
& 79.78 & \underline{176.50} 
& 53.93 & 41.74
& 62.92 & 130.10 \\

MTRouter\citep{zhang2026mtrouter}
& \underline{68.54} & 73.55
& \underline{80.90}  & 216.30
& 60.67 & 51.47
& 71.91 & 160.13 \\

\addlinespace[3pt]

\rowcolor{violet!10}
\multicolumn{9}{c}{\textit{Delegation Strategies (Ours)}} \\

Binary Delegation
& \textbf{71.91} & 59.16
& \textbf{82.02} & 195.63
& \textbf{66.29} & 58.16
& \textbf{79.78} & 168.19 \\



\rowcolor{blue!5}
Self-Delegation
& \textbf{71.91} & \underline{56.96}
& \textbf{82.02} & 181.88
& \underline{64.04} & \underline{42.98}
& \underline{75.28} & \underline{126.53} \\

\rowcolor{blue!5}
\footnotesize\itshape $\Delta$ vs. Vanilla LLM
& \footnotesize\itshape -2.25\% & \footnotesize\itshape -36.93\%
& \footnotesize\itshape +0.00\% & \footnotesize\itshape --34.98\%
& \footnotesize\itshape -3.38\% & \footnotesize\itshape --33.65\%
& \footnotesize\itshape +0.00\% & \footnotesize\itshape --32.60\%\\

\midrule
\multicolumn{9}{c}{\large\textit{\textbf{SWE-ReBench}}} \\


Vanilla LLM
& 56.36 & 200.50
& 63.64 & 598.08
& 60.00 & 199.06
& 69.09 & 584.70 \\

Random Delegation
& 50.00 & 138.38
& \underline{62.73}  & 455.54
& 50.90 & 132.17
& 60.00 & 398.70 \\

RSD\citep{liao2025reward}
& 39.03  & \underline{98.49} 
& 52.73 & \underline{313.09}
& 46.36 & 155.26
& \underline{64.55} & 465.72 \\

STEER\citep{lee2026confidence}
& 48.18  & 125.90
& 60.00  & 383.63
& \underline{51.82} & 141.01
& \textbf{66.36} & 431.38 \\

MTRouter\citep{zhang2026mtrouter}
& 48.18 & 180.11
& \underline{62.73} & 536.45
& 49.09 & 175.20
& 60.00 & 540.90 \\

\addlinespace[3pt]

\rowcolor{violet!10}
\multicolumn{9}{c}{\textit{Delegation Strategies (Ours)}} \\

Binary Delegation
& \underline{55.45} & 127.20
& \textbf{66.36} & 372.60
& 47.27 & \underline{127.63}
& 60.91 & \underline{357.07} \\



\rowcolor{blue!5}
Self-Delegation
& \textbf{58.18} & \textbf{97.80}
& \textbf{66.36} & \textbf{283.02}
& \textbf{57.27} & \textbf{74.28}
& \textbf{66.36} & \textbf{221.39} \\

\rowcolor{blue!5}
\footnotesize\itshape $\Delta$ vs. Vanilla LLM
& \footnotesize\itshape +1.82\% & \footnotesize\itshape -51.22\%
& \footnotesize\itshape +2.72\% & \footnotesize\itshape --52.68\%
& \footnotesize\itshape -2.73\% & \footnotesize\itshape --62.68\%
& \footnotesize\itshape -2.73\% & \footnotesize\itshape --62.14\%\\

\bottomrule
\end{tabular}
}
\end{table*}
We evaluate our delegation strategies against several multi-turn baselines, including LLM-only, random delegation, RSD~\citep{liao2025reward}, STEER~\citep{lee2026confidence}, and MTRouter~\citep{zhang2026mtrouter}. We use Qwen3.5-9B as the SLM and GPT-5.5 (xhigh) or Claude Opus 4.8 via Bedrock\footnote{We observe slightly lower performance for Opus 4.8 via Bedrock due to configuration differences.} as the cloud LLM, and evaluate on Terminal-Bench 2.1 and SWE-ReBench with a maximum of 80 turns. We report Pass@1, Pass@3, and inference cost in~\autoref{tab:main_results}. Cost is computed from cloud LLM API usage based on input and output token pricing; local SLM computation is excluded, while judge API cost is included for binary delegation. Our methods consistently achieve better performance--cost trade-offs than the baselines. In particular, self-delegation reduces Pass@3 cost by 53\%/63\% on SWE-ReBench and 35\%/33\% on Terminal-Bench 2.1 with GPT-5.5/Opus 4.8, respectively, while maintaining performance comparable to LLM-only. This trend is consistent across both cloud LLMs and benchmarks, suggesting that the gains are not tied to a particular model or task distribution. Moreover, the substantial cost reduction indicates that only a fraction of the trajectory requires access to the stronger cloud model, while the remaining steps can be effectively handled by the local SLM. The advantage is more pronounced under Pass@3: with Opus 4.8 on Terminal-Bench 2.1, self-delegation trails LLM-only in Pass@1 but matches it in Pass@3. We hypothesize that SLM--LLM collaboration mitigates overthinking and increases exploration diversity across trajectories. We further study generalization in~\autoref{general} across different harnesses, SLMs and judge models.

\enlargethispage{\baselineskip}

\subsection{Effect of Context Passing on SLM/LLM Collaboration}
\label{context_exp}

\noindent \textbf{Context Provided to LLM.} In this section, we evaluate the context-passing strategies from the SLM to the LLM introduced in~\autoref{what_judge}. For the recent-$k$-steps strategy, we set $k=3$. As shown in the results, passing the full context achieves the best performance but incurs the highest inference cost. We attribute this to long-horizon tasks, where critical steps delegated to the LLM often require more complete historical information. Full context minimizes information loss during the SLM-to-LLM transition. This is further supported by the \textit{no SLM reasoning} strategy, which preserves nearly the full interaction history while excluding the SLM's reasoning, achieving the second-best performance at a lower cost. These results suggest that effective SLM--LLM collaboration requires not only deciding \textit{when} to switch models, but also \textit{what} information to transfer. Preserving sufficient context during model transitions is critical, and context should therefore not be compressed indiscriminately.
\begin{figure*}[htbp]
    \centering
    \includegraphics[width=\textwidth]{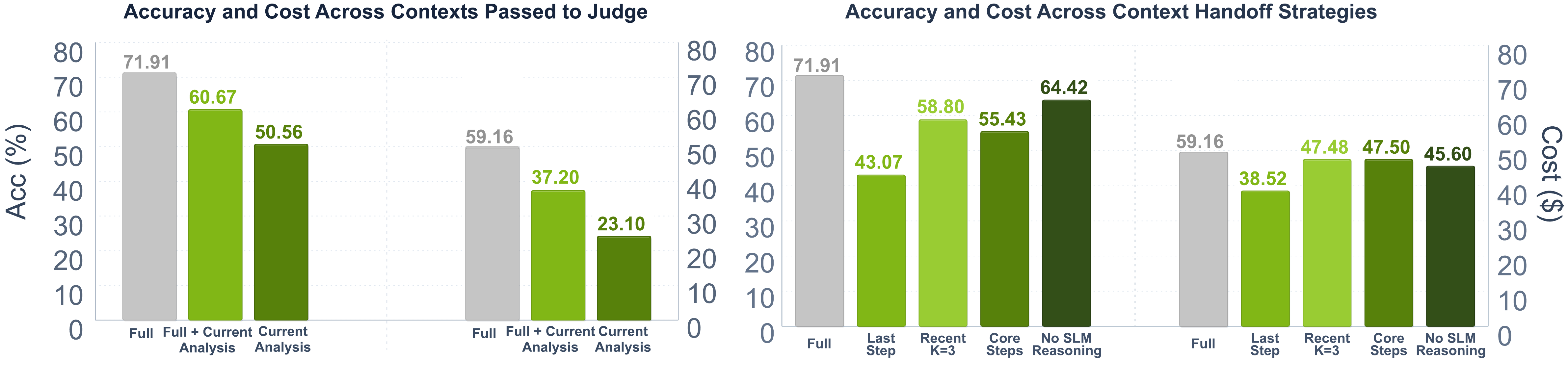}
    \caption{The left panel shows the accuracy and inference cost under three different context configurations provided to the judge. The right panel shows the accuracy and inference cost under five different context configurations passed from the SLM to the LLM.}
    \label{fig:judge-context}
\end{figure*}

\begin{figure*}[htbp]
    \centering
    \includegraphics[width=\textwidth]{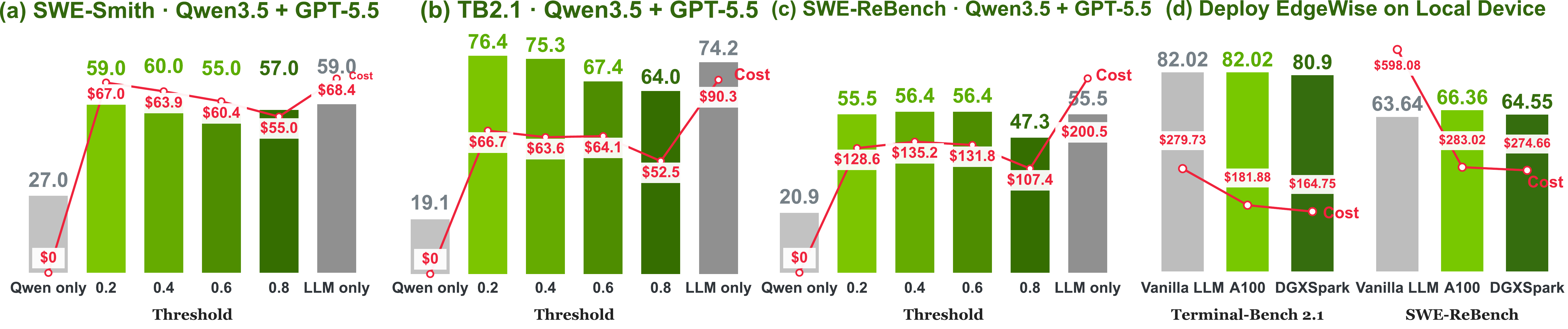}
    \caption{The left 3 figures shows the performance and cost on different thresholds on binary delegation. The right figure shows the comparison between accuracy and inference cost on DGXSpark and A100.}
    \label{fig:threshold-dgxspark}
\end{figure*}

\noindent \textbf{Context Provided to the Judge.} In this section, we evaluate the three judge context configurations introduced in second part of~\autoref{what_judge}. The results are shown in the left panel of~\autoref{fig:judge-context}. Among the three configurations, Full History achieves the best performance. We attribute this advantage to the complete trajectory information available to the judge, which provides a more comprehensive view of previous actions, observations, and task progress. In contrast, incorporating the SLM's current analysis may introduce additional bias when its interpretation is inaccurate, potentially causing the judge to miss critical steps that require LLM intervention. Although this reduces LLM usage and inference cost, it also leads to lower task performance. These results suggest that the complete trajectory provides a more reliable context for delegation decisions.

\subsection{Practical Edge--Cloud Collaboration}
\noindent \textbf{Budget-Aware Cloud Delegation.} Following~\autoref{how_fre}, we introduce a threshold into the external-judge-based method to explicitly control the delegation budget. We first split a validation set from SWE-Smith and evaluate thresholds of \(0.2, 0.4, 0.6, 0.8\). As shown in~\autoref{fig:threshold-dgxspark}(a), thresholds around \(0.2\)--\(0.4\) achieve better performance--cost trade-offs on the validation set. Based on this observation, we further evaluate the method on Terminal-Bench 2.1 and SWE-ReBench with different LLMs, as shown in Figures (b) and (c). The results exhibit a similar trend, where thresholds around \(0.2\)--\(0.4\) maintain strong task performance while substantially reducing inference cost compared with the LLM-only baseline. We further analyze the relationship between the threshold and delegation budget in~\autoref{cloud_budget}.

\noindent \textbf{Real-World Edge Deployment.} In this section, we deploy \sys on a real-world edge device to evaluate its practical feasibility. Specifically, our self-delegation model is deployed on an NVIDIA DGX Spark as the primary solver, selectively delegating steps to a cloud LLM via API when stronger capability is needed. We follow the same experimental configuration as in the previous evaluation. As shown in~\autoref{fig:threshold-dgxspark}(d), the edge deployment achieves Pass@3 performance comparable to our A100-based evaluation while reducing inference cost by approximately 50\% compared with the vanilla LLM. These results demonstrate that \sys preserves its performance-cost benefits in a practical edge--cloud setting. Additional details are provided in~\autoref{dgxspark}.

\noindent \textbf{Real-World Edge Deployment.} In this section, we deploy \sys on a real-world edge device to evaluate its practical feasibility. Specifically, we deploy our self-delegation model on an NVIDIA DGX Spark, which serves as the primary solver throughout task execution. When the local model determines that a step requires stronger intelligence, it delegates the step to a cloud LLM via an API request and subsequently continues execution locally. We follow the same experimental configuration as in the previous evaluation. As shown in~\autoref{fig:threshold-dgxspark}(d), the edge deployment achieves Pass@3 performance comparable to our A100-based evaluation while reducing inference cost by approximately 50\% compared to vanilla model. These results demonstrate that \sys maintains its performance--cost benefits when deployed in a practical edge--cloud setting.
\section{Related Works}
\label{sec:related}

\subsection{Adaptive Model Collaboration}

With the increasing inference cost of large language models, recent studies have explored model routing to reduce computational cost while maintaining model performance. Early methods mainly focus on query-level routing, where the router selects a model before solving the query~\citep{feng2025graphrouter,ong2025routellm,zhao2024eagle}. DiSRouter~\citep{zheng2026disrouter} further allows candidate LLMs to estimate whether they are suitable for solving the current query. Beyond deciding which model to use, R2-Router~\citep{xue2026r2} jointly determines the model and its output-length budget. However, query-level routing makes the routing decision before solving the problem, while the initial query alone may not accurately reflect the actual difficulty encountered during problem solving. This motivates cascade routing, which dynamically determines whether a stronger model is needed based on the output of a smaller model~\citep{aggarwal2024automix,ramirez2024optimising}. Conformal Cascade~\citep{dou2026conformal} uses conformal prediction to estimate the reliability of the small model's answer and escalates to a larger model when the answer is considered unreliable. RLCascadeRouter~\citep{huang2026rlcascaderouter} formulates model invocation as an MDP and uses reinforcement learning to optimize the performance-cost trade-off.

Beyond response-level cascading, other studies explore fine-grained routing during generation. CITER~\citep{zheng2025citer} dynamically determines whether the small or large model should generate the next tokens. R2R~\citep{fu2026r2r} uses a router to identify reasoning-divergent positions and invokes the large model only at these critical positions. While the above methods mainly focus on reasoning generation, routing becomes more challenging in agentic and multi-step tasks, where models need to handle new observations at each step. Router-R1~\citep{zhang2026router} trains a router to dynamically invoke different models from a model pool for multi-hop QA. MasRouter~\citep{yue2025masrouter} trains a router to determine the collaboration mode among multiple agents and allocate an LLM to each agent. MERA~\citep{yao2026mera} assigns easier model invocations to smaller models and falls back to larger models for more difficult ones, while further using successful large-model invocations to train the smaller model.

Despite this progress, existing approaches still have several limitations. First, long-horizon agent tasks, such as coding and deep-research tasks, can involve complex and lengthy trajectories, and existing routing approaches may not scale effectively to such settings. Second, existing methods typically rely on an additional router that selects models from a predefined model pool. Such routers are therefore tied to the specific candidate models considered during routing, making it difficult to generalize when the underlying models are updated or replaced.
\section{Conclusion}

In this work, we present \sys, an efficient agent framework that enables collaboration between local SLMs and stronger cloud models for long-horizon tasks. Our analysis shows that stronger intelligence is primarily needed at a small number of critical steps. Based on this observation, \sys selectively delegates these steps to the cloud model while retaining the remaining execution locally. Across agentic coding benchmarks, \sys substantially reduces cloud-model usage while maintaining competitive task performance, demonstrating the potential of step-level model collaboration for building efficient and practical agent systems.

{
  \small
  \bibliographystyle{unsrt}
  \bibliography{bib}
}

\appendix
\clearpage
\appendix

\section{Limitations}

In this work, we primarily focus on long-horizon coding tasks and do not evaluate our framework on other types of agent tasks, such as deep research or computer-use agents (CUA). While we expect similar opportunities for selective edge--cloud collaboration to exist in these settings, further experiments are needed to validate this hypothesis. We leave the extension to broader long-horizon agent tasks for future work.

\section{Ablation Study}
\subsection{Comparison between SFT model and RL model}
\label{sft-rl}

In this section, we compare our final self-delegation model with the model obtained after the first-stage SFT. As shown in~\autoref{tab:sft}, the SFT-only model achieves substantially lower Pass@1 on Terminal-Bench 2.1. This suggests that SFT primarily teaches the model how to perform delegation and reduce reliance on the LLM, but does not provide sufficient supervision for identifying \textit{when} delegation is necessary. In particular, the SFT labels serve mainly as format supervision and do not explicitly distinguish critical steps from routine ones. As a result, the model may reduce cloud usage without consistently allocating the LLM to the steps where stronger intelligence is most beneficial. In contrast, our subsequent RL stage introduces fine-grained credit assignment for individual delegation decisions, enabling the model to better identify critical steps while maintaining efficient LLM usage. This comparison highlights the importance of the RL stage for learning an effective self-delegation policy rather than simply learning the delegation action.

\begin{table}[htbp]
\centering
\caption{The pass@1 and cost of SFT-model and Self-delegation model.}
\label{tab:sft}
\setlength{\tabcolsep}{10pt}
\renewcommand{\arraystretch}{1.15}
\begin{tabular}{lcc}
\toprule
& Pass@1 & Cost\\
\midrule
SFT-Model                 & 57.30\% & \$ 53.22 \\
Self-Delegation           & 71.91\% & \$56.96 \\

\bottomrule
\end{tabular}
\end{table}

\subsection{The Effectiveness of Token Advantage}
\label{token_adv_good}

To demonstrate the effectiveness of our step-level RL algorithm, we conduct an ablation study on the proposed token-level advantage. Specifically, we use the same reward function as described in~\autoref{rl_method}, while removing the token-level advantage and training the model solely with the standard GRPO advantage. We evaluate both variants on Terminal-Bench 2.1 using GPT-5.5 xhigh as the LLM and report the Pass@1 performance and delegation statistics in~\autoref{tab:ablation}.

The results show that incorporating the token-level advantage not only improves task performance but also leads to more effective delegation behavior. With token-level advantage, 43\% of delegated steps correspond to \textit{write/edit} operations, which we consider more critical steps that typically require stronger reasoning capability, compared with only 29\% without token-level advantage. In contrast, the variant without token-level advantage delegates substantially more \textit{read/search} operations to the LLM. This suggests that fine-grained credit assignment encourages a more selective allocation of cloud-model usage across different types of steps. Moreover, the model trained with token-level advantage achieves better performance with a noticeably lower average number of delegations, indicating that fine-grained credit assignment enables the model to allocate expensive LLM calls more selectively to critical steps and this critical step may harder so need more reasoning tokens which lead higher cost.

\begin{table}[htbp]
\centering
\caption{Ablation study of the token advantage during the RL training stage.}
\label{tab:ablation}
\renewcommand{\arraystretch}{1.15}

\begin{adjustbox}{max width=\columnwidth}
\begin{tabular}{lcc}
\toprule
& w/ Token Advantage & w/o Token Advantage \\
\midrule
\textbf{Acc.}                       & 71.91\% & 65.17\% \\
\textbf{Cost}                       & \$56.96 & \$55.05 \\
\textbf{Write / Edit Rate}          & 42.82\% & 29.20\% \\
\textbf{Read / Search Rate}         & 32.11\% & 40.73\% \\
\textbf{Avg. Delegations / Task}    & 5.80    & 8.97    \\
\bottomrule
\end{tabular}
\end{adjustbox}

\end{table}

\subsection{Effect of SLM Capability on Self-Delegation}
\label{whether-our}

In this section, we investigate whether the effectiveness of self-delegation can be attributed simply to improved SLM capability rather than our delegation-specific training. As a control, we first train Qwen3.5-9B with standard GRPO on the same training data to improve its task-solving capability. We then apply the same SFT stage to teach the strengthened model the delegation format and action, but without introducing our step-level delegation reward. The resulting model therefore has improved task-solving capability and can perform delegation, but does not receive explicit supervision on \textit{when} to delegate. As shown in~\autoref{tab:trained_slm_only}, this stronger SLM with delegation still underperforms our self-delegation method. This suggests that stronger task-solving capability alone, even when combined with the ability to delegate, is insufficient to learn an effective delegation policy. Instead, our step-level credit assignment plays an important role in teaching the model where cloud assistance is most beneficial This observation is further supported by the random-delegation experiment in~\autoref{fig:random-selection}. Using the original Qwen3.5-9B without additional training, different placements of cloud-model calls lead to substantial performance variation, with some allocations achieving performance comparable to the vanilla LLM while using substantially fewer cloud calls. Together, these results suggest that effective SLM--LLM collaboration depends not only on the capability of the local SLM, but critically on learning \textit{when} to delegate.

\begin{table}[htbp]
\centering
\caption{Comparison of using GRPO train Qwen3.5-9B and self-delegation model.}
\label{tab:trained_slm_only}
\setlength{\tabcolsep}{10pt}
\begin{adjustbox}{max width=\columnwidth}
\renewcommand{\arraystretch}{1.15}
\begin{tabular}{lcc}
\toprule
\textbf{Model} & \textbf{Cloud LLM} & \textbf{Pass@1 (\%)} \\
\midrule
Vanilla Qwen3.5-9B                  & \ding{55}  & 19.10    \\
Vanilla + GRPO + SFT      & \ding{51}  & 32.58   \\
Self-Delegation   & \ding{51} & 71.91 \\
\bottomrule
\end{tabular}
\end{adjustbox}
\end{table}

\section{More Experiments}
\label{general}

\subsection{Generalization Across Different SLMs}

In this section, we evaluate our framework with a different SLM to examine its generalizability across local models. Specifically, we use TMax-9B as the SLM and GPT-5.5 xhigh as the LLM, and evaluate Pass@1 on Terminal-Bench. As shown in~\autoref{tab:tmax}, our framework exhibits a similar trend to that observed with Qwen3.5-9B. With TMax-9B as the local model, our method achieves the same Pass@1 as GPT-5.5 while reducing inference cost by 30.6\%.

\begin{table}[htbp]
\centering
\caption{The Pass@1 and cost of using Tmax-9B as the SLM in our binary delegation framework.}
\label{tab:judge_alter}
\renewcommand{\arraystretch}{1.15}

\begin{adjustbox}{max width=\columnwidth}
\begin{tabular}{lccc}
\toprule
& Pass@1 & Judge Cost & LLM Cost \\
\midrule
Qwen3.5-9B  & 50.56\% & \$0.00 & \$62.11 \\
Qwen3.8-27B & 70.79\% & \$0.00 & \$75.74 \\
GPT-5.4-mini & 71.91\% & \$5.77 & \$53.39 \\
\bottomrule
\end{tabular}
\end{adjustbox}

\end{table}

\subsection{The alteration of Judge Model}
In this section, we investigate the impact of using different judge models in the binary delegation method. We consider three judges with different model capacities: Qwen3.5-9B (the SLM itself), Qwen3.8-27B, and GPT-5.4-mini. We keep Qwen3.5-9B as the SLM and GPT-5.5 (xhigh) as the LLM, and evaluate Pass\@1 on Terminal-Bench 2.1. The results are shown in~\autoref{tab:judge_alter}. We observe that using the SLM itself as the judge leads to relatively poor performance. However, as the capability of the judge model increases, Qwen3.8-27B and GPT-5.4-mini achieve comparable Pass\@1 performance. Although using GPT-5.4-mini incurs a lower inference cost than Qwen3.8-27B, the similar performance achieved by the two models demonstrates that our framework is not tied to a specific judge model. In particular, an open-source model such as Qwen3.8-27B can serve as an effective alternative to a proprietary judge while maintaining comparable task performance.

\begin{table}[htbp]
\centering
\caption{The pass@1 and cost of using Tmax-9B as SLM of our binary delegation method in the framework.}
\label{tab:tmax}
\setlength{\tabcolsep}{10pt}
\renewcommand{\arraystretch}{1.15}
\begin{tabular}{lcc}
\toprule
& Pass@1 & Cost\\
\midrule
Tmax-9B                  & 28.09\% & \$ 0 \\
Binary Delegation             & 74.16\% & \$62.64 \\
GPT-5.5(xhigh)          & 74.16\% & \$90.31 \\
\bottomrule
\end{tabular}
\end{table}

\begin{figure}[htbp]
    \centering
    \includegraphics[width=\columnwidth]{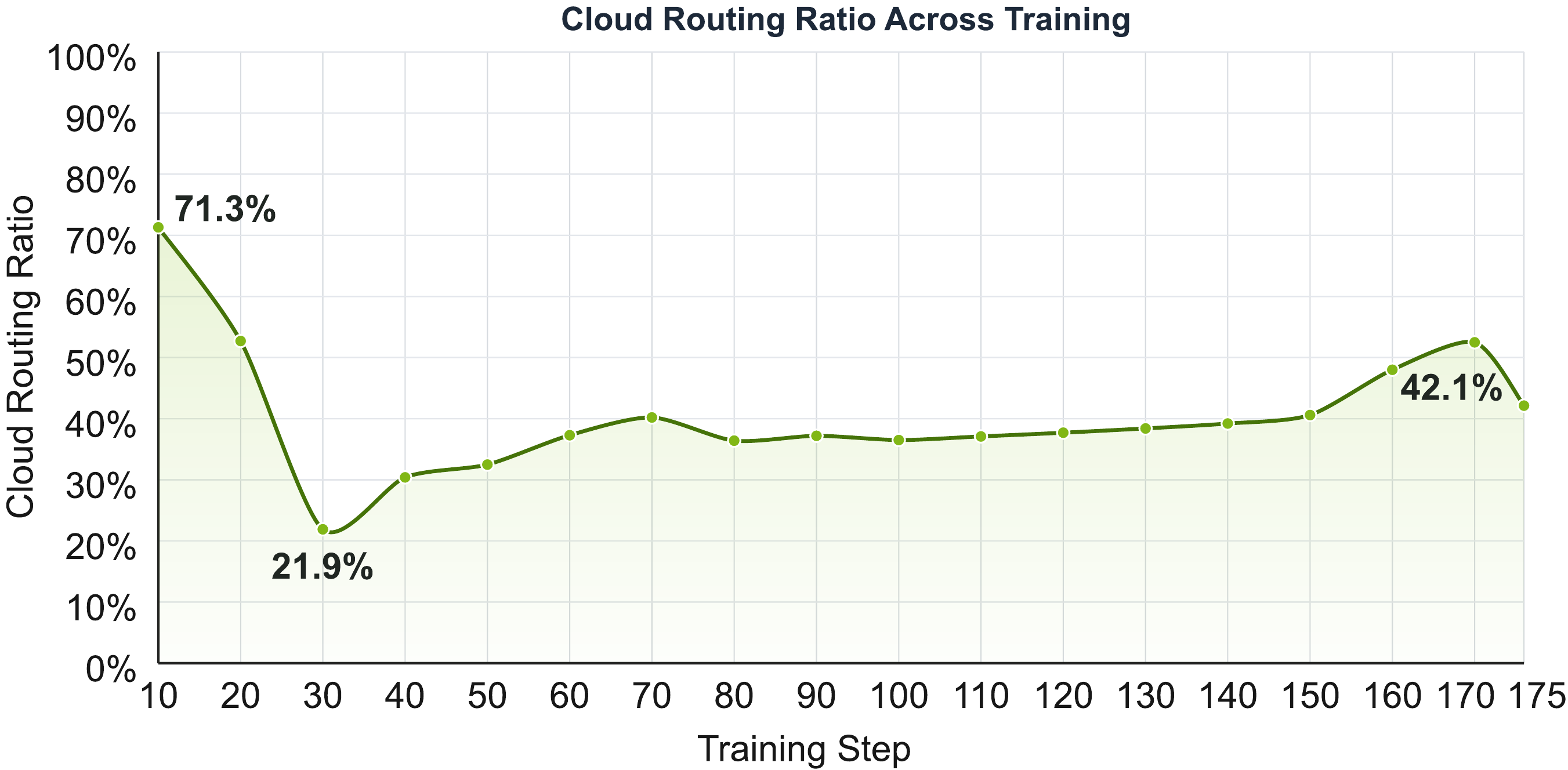}
    \caption{LLM delegation rate in answer trajectories throughout the training process.}
    \label{fig:cloud_rate}
\end{figure}

\subsection{LLM rate in the final answer trajectory}
\label{call_time}

In this section, we examine the LLM delegation rate across different training checkpoints to investigate how the model's tendency to delegate to the cloud evolves during training. The results are shown in~\autoref{fig:cloud_rate}. At the beginning of training, immediately after SFT, the model has learned the basic delegation behavior but has not yet learned when delegation is appropriate. Consequently, a large proportion of steps are delegated to the cloud LLM. As training progresses, the delegation rate decreases substantially, indicating that the penalty on LLM delegation in our reward function effectively encourages the model to reduce unnecessary cloud calls. After reaching its minimum, the delegation rate gradually increases again. We attribute this trend to our token-level advantage, which provides fine-grained supervision and enables the model to progressively learn when delegation to the LLM is beneficial, rather than simply minimizing the number of delegations. Eventually, the delegation rate stabilizes and fluctuates around 40\%, at which point we select the final checkpoint for evaluation.
\begin{figure*}[t]
    \centering
    \includegraphics[width=\textwidth]{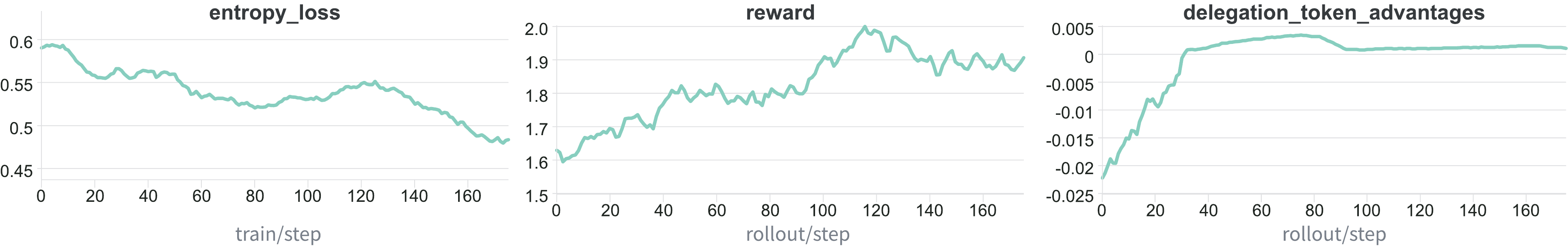}
    \caption{Learning curve during the training. From left to right, the three panels illustrate entropy loss, reward and the advantage of judge tokens.}
    \label{fig:learning-curve}
\end{figure*}

\subsection{Generalization of Self-Delegation}

In this section, we evaluate the generalizability of the checkpoint trained using our self-delegation method. We use GPT-5.5 with xhigh reasoning effort as the cloud LLM and evaluate the checkpoint on Terminal-Bench 2.1 using the OpenHands, $\pi$, and mini-SWE-agent harnesses. As shown in~\autoref{tab:collaboration}, although our self-delegation model is trained exclusively with the Terminus2 harness, it maintains Pass@3 performance comparable to the vanilla LLM when transferred to all three unseen harnesses, while reducing inference costs by at least 25\%. These results demonstrate that our method generalizes effectively across different agent harnesses.


\begin{table}[t]
\centering
\caption{Pass@3 using different harness on Terminal-bench 2.1 .}
\label{tab:collaboration}

\footnotesize
\setlength{\tabcolsep}{3.5pt}
\renewcommand{\arraystretch}{1.08}

\resizebox{\columnwidth}{!}{%
\begin{tabular}{llcc}
\toprule
\multirow{2}{*}{Harness}
& \multirow{2}{*}{Method}
& \multicolumn{2}{c}{Qwen3.5 + GPT-5.5 (xhigh)} \\
\cmidrule(lr){3-4}
& & Pass@3 & Cost \\
\midrule

\multirow{4}{*}{Terminus-2}
& Vanilla SLM
& 29.10
& 0.00
\\
& Vanilla LLM
& 82.02
& 279.73
\\
& Self-Delegation
& 82.02
& 181.88
\\
\rowcolor{blue!5}
& \footnotesize\itshape $\Delta$ vs.\ LLM
& \footnotesize\itshape +0.00\%
& \footnotesize\itshape -34.89\%
\\

\midrule

\multirow{4}{*}{OpenHands}
& Vanilla SLM
& 20.22
& 0.00
\\
& Vanilla LLM
& 78.65
& 267.82
\\
& Self-Delegation
& 75.28
& 200.84
\\
\rowcolor{blue!5}
& \footnotesize\itshape $\Delta$ vs.\ LLM
& \footnotesize\itshape -3.37\%
& \footnotesize\itshape - 25.09\%
\\

\midrule

\multirow{4}{*}{Pi}
& Vanilla SLM
& 21.35
& 0.00
\\
& Vanilla LLM
& 79.78
& 298.87
\\
& Self-Delegation
& 76.40
& 221.35
\\
\rowcolor{blue!5}
& \footnotesize\itshape $\Delta$ vs.\ LLM
& \footnotesize\itshape -3.38\%
& \footnotesize\itshape -25.94\%
\\

\midrule

\multirow{4}{*}{Mini-SWE-Agent}
& Vanilla SLM
& 22.47
& 0.00
\\
& Vanilla LLM
& 83.15
& 303.88
\\
& Self-Delegation
& 80.90
& 163.12
\\
\rowcolor{blue!5}
& \footnotesize\itshape $\Delta$ vs.\ LLM
& \footnotesize\itshape -2.25\%
& \footnotesize\itshape -46.32\%
\\

\bottomrule
\end{tabular}%
}

\end{table}

\section{Learning Curve}
\label{learning_curve}

In this section, we show the learning curve of some crucial metrics during the training. The details in~\autoref{fig:learning-curve}. We use a running average to smooth the curves. In addition to the standard training dynamics of entropy and reward, we focus on the rightmost panel, which shows the evolution of the advantage assigned to delegation tokens. At the early stage of training, the average advantage is negative, indicating that many delegation decisions are inconsistent with the critic's feedback and therefore receive corrective updates. As training progresses, the advantage increases and eventually converges toward zero, suggesting that the model gradually adapts its delegation behavior and fewer delegation tokens require strong corrective updates. This trend provides evidence that our step-level credit assignment effectively guides the model toward more appropriate delegation decisions.

\begin{figure}[htbp]
    \centering
    \includegraphics[width=\columnwidth]{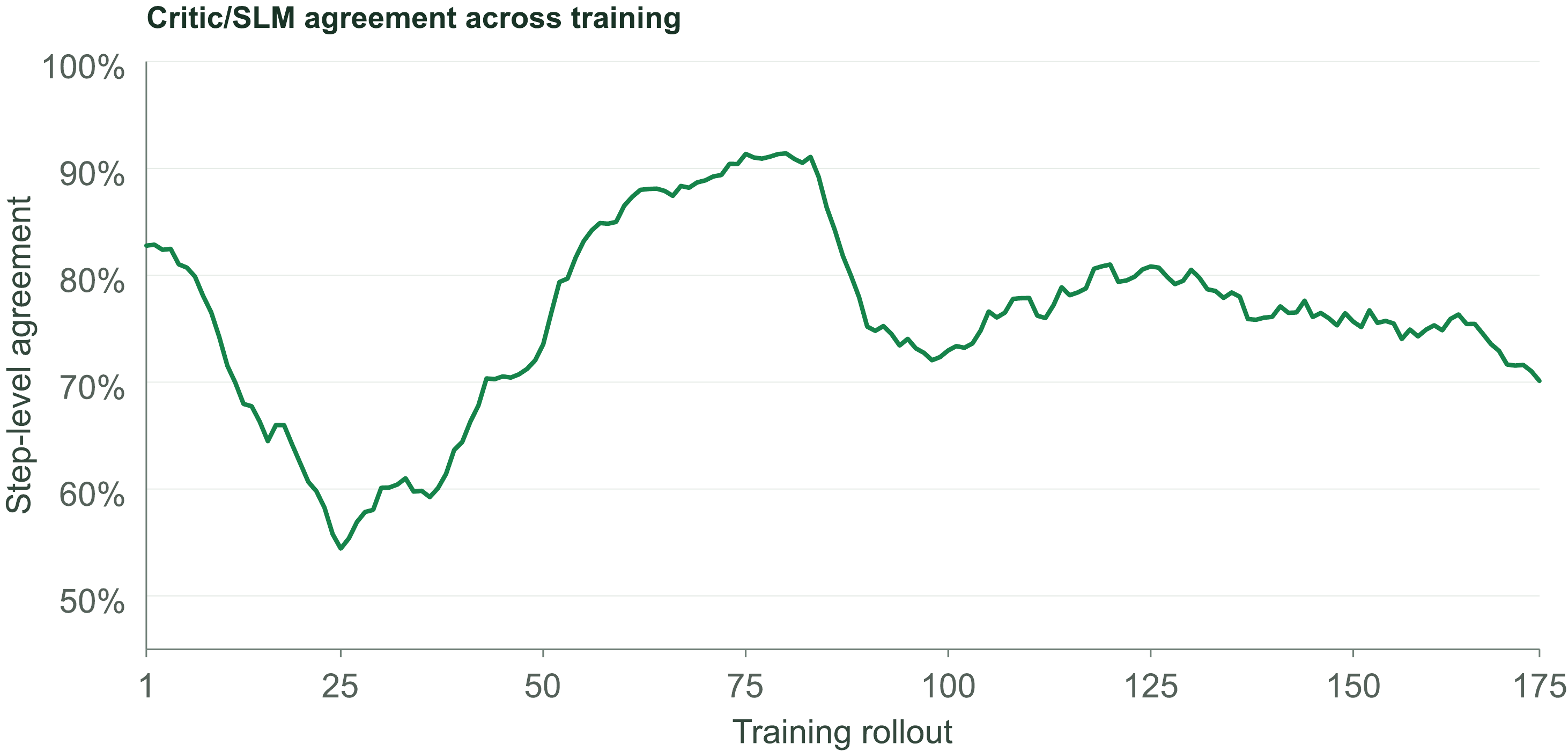}
    \caption{The critic and SLM agreement with the training steps.}
    \label{fig:critic-agree}
    \vspace{-5pt}
\end{figure}

We give futher analyze about the agreement of critic and SLM, the results in~\autoref{fig:critic-agree}. At the beginning of RL training, the critic agreement first decreases. This is expected because the SFT stage primarily teaches the model the delegation format rather than \textit{when} to delegate. Once RL optimization begins, the policy moves away from its SFT initialization under the trajectory-level task and cost rewards, while its delegation behavior has not yet been well aligned with the step-level supervision. Consequently, many delegation decisions receive negative advantages, as reflected by the strongly negative delegation-token advantage at the beginning of training. As training progresses, our step-level credit assignment begins to correct these decisions: inappropriate delegation decisions receive negative feedback, while appropriate ones are reinforced. Correspondingly, the delegation-token advantage increases and the critic agreement recovers substantially, reaching over 90\% at its peak. These consistent trends provide evidence that the step-level reward effectively teaches the model \textit{when} to delegate, rather than merely reducing the overall number of LLM calls. In later training, the agreement stabilizes below its peak and the delegation advantage approaches zero, as the policy jointly balances task success, inference cost, and critic guidance instead of simply imitating the critic.

\section{Distribution of LLM-Handle Steps}
\label{distribution}

\begin{figure*}[t]
    \centering
    \includegraphics[width=\textwidth]{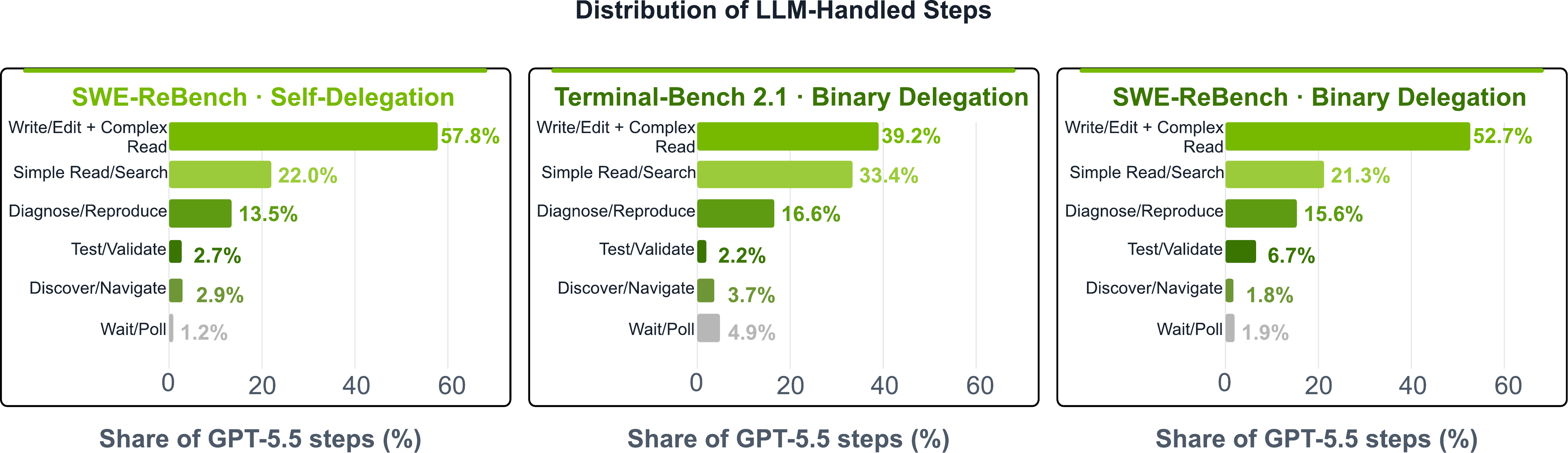}
    \caption{Distribution of step types handled by the LLM on Terminal-Bench 2.1 and SWE-ReBench.}
    \label{fig:cloud_rate}
\end{figure*}

In this section, we further analyze the types of steps handled by GPT-5.5 under both self-delegation and binary delegation across the two benchmarks. The results reveal a consistent delegation pattern across different tasks and strategies. In both settings, the LLM is primarily invoked for more complex operations, particularly write/edit steps and complex read/search steps. Write/edit operations often require understanding the existing codebase, identifying the correct modification location, and generating changes that remain consistent with the surrounding implementation, making them more challenging than routine execution steps. Similarly, read/search operations can involve cross-file exploration and dependency tracing, which require reasoning over broader repository context. This pattern is consistent with our earlier analysis of critical steps: stronger model capabilities provide greater benefits on steps that require more complex reasoning and cross-file understanding. Importantly, the LLM is not invoked uniformly throughout the trajectory; instead, both delegation mechanisms concentrate cloud-model usage on these more challenging operations while leaving simpler steps to the local SLM. This further suggests that our methods can effectively identify where stronger intelligence is most valuable and selectively allocate the LLM to those steps.

\section{The Details of Cloud Budget in the Budget-Aware Cloud Delegation}
\label{cloud_budget}

\begin{figure}[htbp]
    \centering
    \includegraphics[width=\columnwidth]{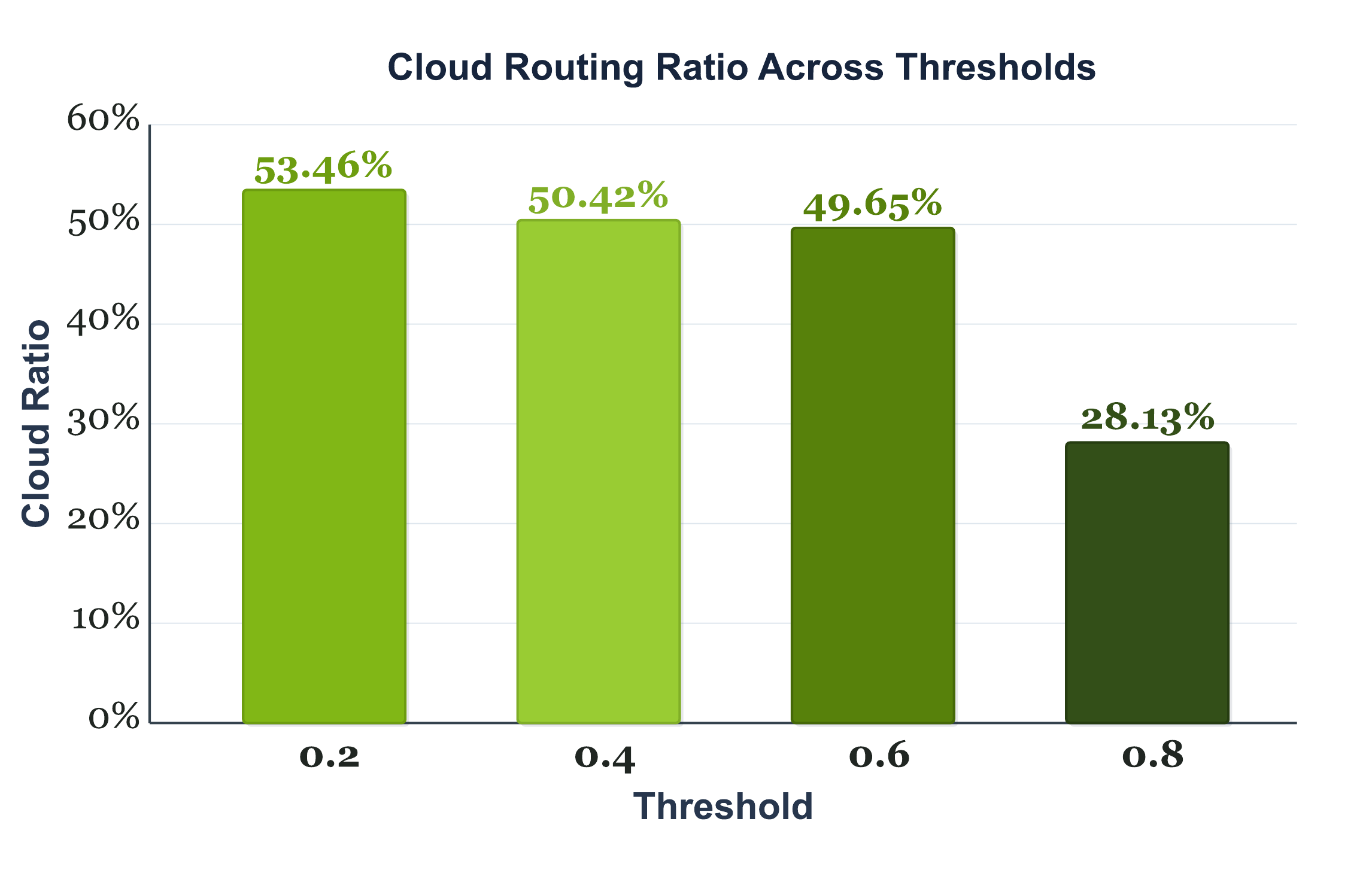}
    \caption{The cloud rate under different thresholds.}
    \label{fig:threshold-cloud-rate}
    \vspace{-10pt}
\end{figure}

In this section, we further investigate how the delegation threshold affects the actual cloud-model usage. We use Qwen3.5-9B as the local SLM and GPT-5.5 (xhigh) as the cloud LLM, and evaluate the delegation behavior under different threshold values. For each threshold, we measure the cloud rate, defined as the proportion of agent steps that are delegated to the cloud LLM during task execution. As shown in~\autoref{fig:threshold-cloud-rate}, the cloud rate consistently decreases as the delegation threshold increases. This is expected because a higher threshold requires a higher delegation score before invoking the cloud LLM, causing more steps to remain on the local SLM. These results establish an empirical relationship between the delegation threshold and actual cloud-model usage, showing that the threshold can serve as an effective control parameter for adjusting the level of cloud involvement. We note that the threshold itself does not directly specify a fixed delegation budget; rather, it indirectly controls cloud usage by changing the criterion for delegation. Therefore, different threshold values correspond to different empirical levels of cloud usage, with larger thresholds generally resulting in fewer cloud-model invocations.

\section{Details of Implementation}
\subsection{Details of Token-level Advantage}
Here, we provide the implementation details of the delegation-token advantage. When both the SLM and the critic determine that the current step should be delegated to the LLM, we assign an advantage of $+1$; when their decisions disagree, we assign an advantage of $-1$. Here, $d_i$ denotes the delegation decision made by the SLM, while $\hat{d}_i$ denotes the decision made by the critic model.

\begin{equation}
A_i^{\mathrm{del}} =
\begin{cases}
+1.0,  & d_i= \hat{d}_i \\
-1.0,  & d_i \neq \hat{d}_i
\end{cases}
\end{equation}

\subsection{Details of Deploying on DGXSpark}
\label{dgxspark}

We deploy Qwen3.5-9B on a single NVIDIA DGX Spark using vLLM 0.19.1 with BF16 precision and single-device inference, setting the GPU memory utilization to 0.65. The deployment reserves approximately 78.7 GB of unified memory at runtime, including 16.8 GiB for model weights and approximately 57.1 GiB available for the KV cache. During active serving, vLLM achieves an average generation throughput of 11.19 tokens/s. On Terminal-Bench 2.1, the average and median end-to-end task completion times are 11.72 and 8.30 minutes, respectively. These measurements provide the deployment and runtime characteristics of EdgeWise on a practical edge device.

\end{document}